\documentclass[runningheads]{llncs}
\usepackage{graphicx}
\usepackage{amsmath}
\usepackage{amssymb}

\usepackage{capt-of}% or \usepackage{caption}
\usepackage{booktabs}
\usepackage{varwidth}

\usepackage[export]{adjustbox}

\begin{document}
\title{Deep Multi-class Adversarial Specularity Removal}
\author{John Lin\inst{1,2} \and
Mohamed El Amine Seddik\inst{1} \and
Mohamed Tamaazousti\inst{1} \and Youssef Tamaazousti\inst{1} \and Adrien Bartoli\inst{2}}
\authorrunning{J. Lin et al.}
\institute{CEA, LIST, Point Courrier 184, Gif-sur-Yvette, F-91191, France \\
\email{\{john.lin, mohamedelamine.seddik, mohamed.tamaazousti\}@cea.fr} \\
\email{youssef.tamaazousti@gmail.com} \and
Institut Pascal - UMR 6602 - CNRS/UCA/CHU, Clermont-Ferrand, France \\
\email{adrien.bartoli@gmail.com} \\}
\maketitle              % typeset the header of the contribution
\begin{abstract}
We propose a novel learning approach, in the form of a fully-convolutional neural network (CNN), which automatically and consistently removes specular highlights from a single image by generating its diffuse component. To train the generative network, we define an adversarial loss on a discriminative network as in the GAN framework and combined it with a content loss. In contrast to existing GAN approaches, we implemented the discriminator to be a multi-class classifier instead of a binary one, to find more constraining features. This helps the network pinpoint the diffuse manifold by providing two more gradient terms. We also rendered a synthetic dataset designed to help the network generalize well. We show that our model performs well across various synthetic and real images and outperforms the state-of-the-art in consistency.

\keywords{Deep learning  \and GAN \and Dichromatic reflection separation \and Specular and diffuse components.}
\end{abstract}

\section{Introduction}
The appearance of an object depends on the way it reflects light. The materials constituting most objects have a dichromatic behaviour: they produce two types of reflection, namely diffuse and specular reflections. Shafer's dichromatic model~\cite{shafer} linearly combines these two terms for the image formation model. The fundamental difference between them is that the diffuse reflection does not change with the viewing direction, while the specular one does. 
The specular reflection is thus largely responsible for the tremendous difficulty of solving for the parameters of an image formation model.
It is then appealing to assume the object's surface to be perfectly diffuse and rule out the specular reflection.
This has been extensively used in various computer vision problems, including SLAM, image segmentation and object detection, to name but a few.
The price to pay however is failure of the methods when the real object's reflection departs, sometimes even slightly, from diffusion.
%
%\begin{figure} [t!]
%\centering
%	\includegraphics [width = \linewidth]{figures/fish.pdf}
%	\label{fish}
%	\caption{Specular and Diffuse separation with our approach. Our method takes a single input image (a) and directly generates its diffuse component (b) without priors on the scene. The specular component (c) is then obtained by subtracting (b) to (a).}
%\end{figure}

Quite naturally, many approaches have been proposed to solve the problem of specular and diffuse separation, and to be applied as a preprocessing to many algorithms. The separation is also relevant in computer graphics since the specular component conveys precious information about the surface material and illumination~\cite{hara2003determining,lin1999estimation}. We can group model-based methods in two categories: multi-image and single-image approaches. Multi-image methods ingeniously use specular reflections' physical properties to find and remove specularities from an image, such as their polarization properties \cite{nayar1997separation,wolff1991constraining} or their dependance to the viewpoint to find matching specular and diffuse pixels from several images \cite{jachnik2012real,lin2002diffuse,lin2001separation}. While obtaining good results, these methods are impratical to use because of the need for multiple images, special equipment (polarizer) or known object geometry. Single-image methods mostly rely on the Dichromatic Reflection Model and the fact that specularities retain the illumination's color to do the separation \cite{an2015fast,klinker1988measurement,shafer,shen2009simple,tan2005separating}. However, the separation problem with a single image being ill-posed because of the ambiguity of the image formation process \cite{adelson1996perception}, they make strong assumptions about the scene such as a single illumination of known color, no saturated pixels and no nonlinearity of the capture device. This obviously hinders the generic applicability of the methods. Therefore, this is still a challenging and open problem.\par
In this paper, we propose a deep learning approach to overcome the limitations in applicability. The idea is that the network will work out the intricate relationships between an image and its diffuse part. Recently, a handful of learning-based methods have been proposed to solve the diffuse and specular separation problem \cite{funke2018generative,meka2018live,shi2017learning}.
Such data-driven approaches reduce the need to find hand-crafted features and priors, which might not even be relevant for the wide diversity of possible scenes~\cite{weiss2001deriving}. An immediate challenge is to find a large scale real dataset, since it is extremely time-consuming to produce one. Therefore, we train our network on synthetic data. We specifically rendered the data to overcome some limitations known to the problem of separation, by including known causes of failure cases in hand-crafted methods. Another challenge of learning approaches is to generalize, all the more difficult when training with synthetic data. To overcome this limitation, we build our work on the fairly recent framework of Generative Adversarial Networks (GAN)~\cite{goodfellow2016nips}, which we adapt to the separation problem. Just as in GANs, we have a generator network, which we call Specularity Removal Network, trained to generate the diffuse image, while the discriminator network is used only for training by determining whether specularities are well removed. The main difference with a classical GAN resides in the discriminator network, which is not a binary classifier but a categorical classifier. By increasing the number of classes, we help the discriminator pinpoint the desired manifold. This allows it to find more discriminative features for the task at hand. It also prevents an unwanted behaviour of the GAN on synthetic data \textit{i.e.} to generate data that look synthetic. Our method takes a single RGB image as input and does not make any assumption about the scene. We show in the results \S\ref{Results} that our framework is more stable than existing methods~\cite{shen2008chromaticity,shi2016real,tan2005separating} for a wide range of images, outperforming them qualitatively and quantitatively.

In summary, our work addresses the aforementioned challenges and makes the following contributions:
    
    \begin{itemize}
        \item A new method of Single-image Specular-Diffuse Separation (SSDS), free from priors on the scene and capable of performing on a wide range of images.
        \item A new multi-class adversarial loss for the problem of SSDS. 
        \item A new synthetic dataset, designed for the task of specular highlights removal.
    \end{itemize}

%------------------------------------------------------------------------
\section{Deep Specularity Removal}
\begin{figure}[h]
\centering
   \includegraphics[width=\linewidth]{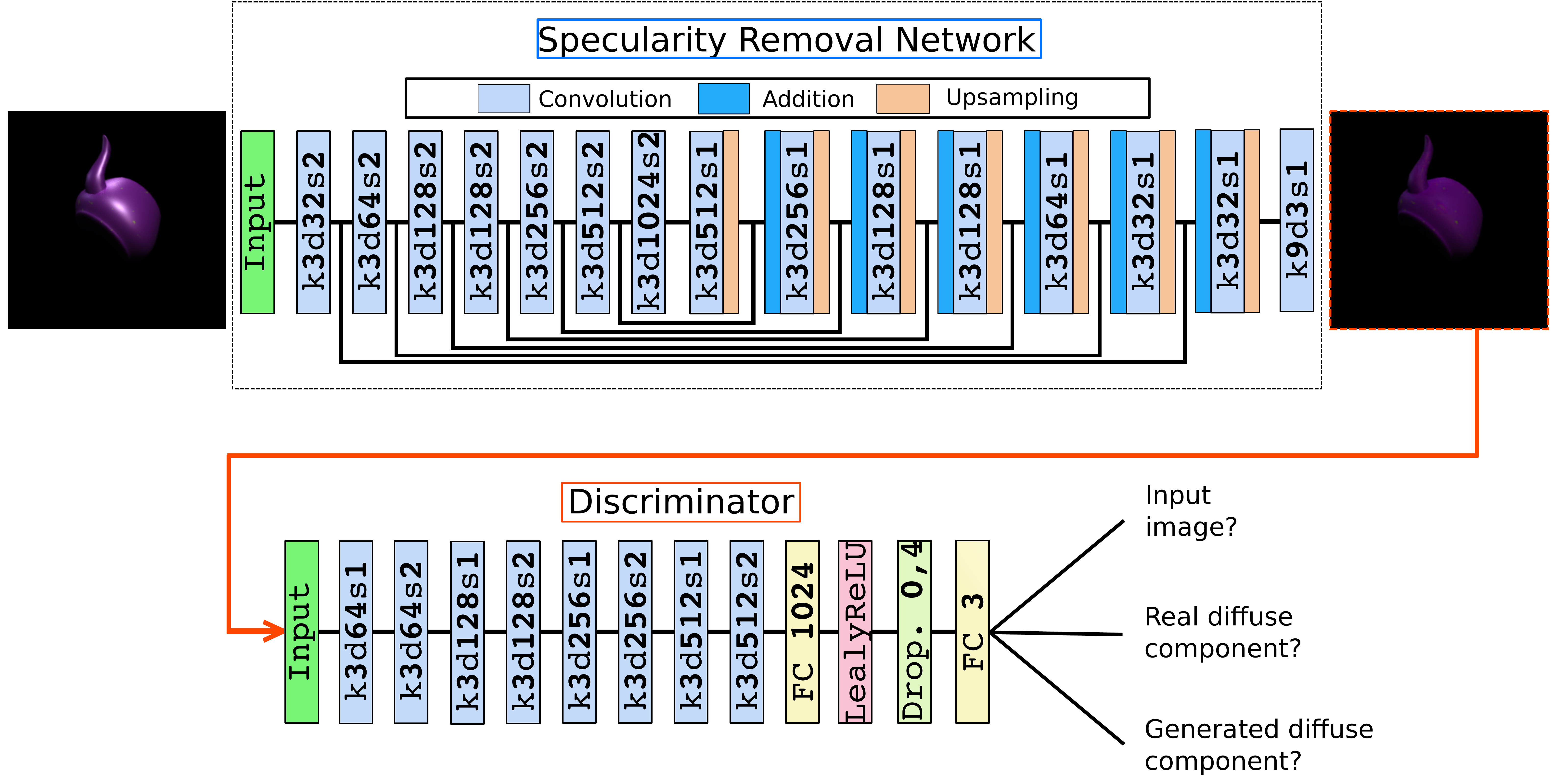}
   \caption{Overview of our architecture. The Specularity Removal Network takes a single image as input and outputs its diffuse component. The discriminator network is used only for training. It takes the entire image as input and is trained to classify an image into three categories: the input image ($I$), the diffuse component ($D$) and the generated diffuse component ($\Hat{D}$); while the generator is trained to fool the discriminator to classify $\Hat{D}$ as $D$.}
   \label{fig:overview}
\end{figure}
Following the dichromatic model~\cite{shafer}, when light hits an object, it is divided in two parts at the surface, at a ratio depending on the material's refraction index. One part, called the specular reflection, is reflected at the surface, in the manner of a mirror reflection. The other part, called the diffuse reflection, penetrates the object and scatters before coming out and being reflected. These two bounced off parts of light then add up, and after integrating all lights coming from the upper hemisphere of the object surface, the image is formed. Integration being a linear operation, we can describe an image $I$ as:
\begin{equation}
    I = S + D,
\end{equation}
where $S$ is the specular image and $D$ the diffuse image.
The problem of Single image Specular and Diffuse Separation (SSDS) consists in estimating the specular $S$ and/or the diffuse $D$ component of a given image $I$, since a consistent estimation of one component suffices to retrieve the other by subtracting from $I$. Thus, we consider the problem of predicting the diffuse component $D$ for which we have more visual cues since there are often purely diffuse pixels in images.

\subsection{Overview}
To solve the ill-posed problem of SSDS, we propose a generator $\mathcal{G}_{\theta_g}$, parametrized by its weights $\theta_g$. $\mathcal{G}_{\theta_g}$ is a feed-forward CNN, which takes $I$ as input and generates $D$. To train $\mathcal{G}_{\theta_g}$, we carefully rendered a realistic and diverse synthetic training set $\mathbb{T}=\{ I_i, D_i\}_{i=1}^N$ of $N$ images and their corresponding ground truth diffuse components. The dataset generation is discussed in \S\ref{sec:Dataset}. Formally, our method boils down to the following optimization problem:
\begin{equation}
\hat{\theta}_g = \arg\min_{\theta_g} \frac{1}{N} \sum_{i=1}^N \ell \left( \mathcal{G}_{\theta_g}(I_i), D_i \right),
\label{eq:final_obj}
\end{equation}
where $\ell$ is our SSDS-specific loss function. $\ell$ is a weighted combination of two loss components. One of these components is a content-loss to drive the learning and the other is an adversarial loss to increase the accuracy of predicted diffuse components. The discriminative network is specifically designed for the task of SSDS, as shown in the results. It is trained to recognize whether or not specularities were well removed from an image, while the specularity removal network is trained to fool them. The discriminator network and the loss are discussed in \S\ref{sec:disc} and \S\ref{sec:loss} respectively. %Figure~\ref{fig:overview} gives an overview of our framework.

\subsection{Synthetic Training Dataset}
\label{sec:Dataset}

Our training set $\mathbb{T}=\{ I_i, D_i\}_{i=1}^N$ consists of $N = 20000$ realistically rendered pairs of images. The generation is automated with a script using Blender and the Cycles engine. Each of the $N$ training frames shows a single object from a set of 8 synthetic 3D models, rendered at the center of the image in a random orientation. We excluded 5 shapes of the training set to also create a test set of 1000 images used in the quantitative evaluation. The diffuse part is rendered with the Lambertian model \cite{lambert1760photometria} and the specular part with the Beckmann distribution of a microfacet model (Glossy BSDF in Blender). The specular roughness is randomly chosen in the range $[0.2, 0.5]$. We set a lower bound to this roughness so that the material does not become highly specular, almost mirror-like. To handle mirror-like surfaces, we would need a different dataset and maybe a different network all-together. Among these $N$ frames, we have four sets of data, each one rendered to overcome a learning limitation.

%\subsection*{Random texture objects}
\subsubsection{Random texture objects.} This first set has 10000 images rendered with four area lights, directed towards the object. The position of the lights as well as their intensity are randomized, to increase the diversity of the scenes. With the same intent, we set a random colored texture to each object. This simulates real life objects which do not always show uniform reflectance and also compels the network to render images of high quality by focusing on details. The lights' color is fixed to near white (two are slightly blue and the others yellow to imitate real lights).\par

%\subsection*{White objects}
\subsubsection{White objects.} With only the first set, the generative network simply learns to remove white pixels from the image as it suffices for the training set. Therefore we add this second set with 4000 images of white objects, otherwise rendered in the same manner as the first set, to help the network grasp the difference between white objects and specularities.\par

%\subsection*{Colored lights}
\subsubsection{Colored lights.} This third set contains 2000 images rendered as the first set but with a random color assigned to each light. This allows us to take into account the real life cases where the lights are not white, although it is not very common (which is why we render only 2000 images for this set). Having different light colors in the scene also ensures that our network does not overfit our dataset by only analyzing white pixels and their intensity.

%\subsection*{Environment maps}
\subsubsection{Environment maps.} Finally, this last set was added to the corpus because in real life specular highlights sometimes spread over a large portion of the object instead of being small and localized. This is due to the inter-reflections that occur in real scenes, whereas our synthesized data only contain one object. To simulate this effect, we render this set with an environment map $E_i$, randomly sampled from a set of 6 High Dynamic Range (HDR) maps.

%\begin{figure}[tb!]
%\centering
%\includegraphics [width = 0.7\linewidth]{figures/dataset.pdf}
%\vspace{-0.15cm}
%\caption{\label{data_ex}
%Four examples of our proposed training dataset. Each column shows a set from those described in Section~\ref{sec:Dataset} (presentation of the examples are in the same order than the description in Section~\ref{sec:Dataset}). Best view in PDF.}
%\end{figure}

\subsection{Specularity Removal Network}
\label{sec:cnn}
The specularity removal network is a CNN with skip connections, inspired by U-Net~\cite{ronneberger2015u}. It takes in a single RGB image, decreases its resolution with strided convolutions before further processing and upsampling it to generate its diffuse counterpart. Figure~\ref{fig:overview} (top) depicts the architecture of our generator network, where k, d and s respectively mean the kernel size, the depth and stride of the convolutions.We use the ReLU activation function at each convolutional layer. The network was trained with images of size $256\times256$, but can then be applied to images of any size as it is fully convolutionnal. \par

\subsection{Discriminator Network}
\label{sec:disc}
In classical GANs, the discriminator network is a binary classifier, trained to recognize real images from fake ones. Then, the generator is trained to fool it, which results in high perceptual quality image generation, close to the data's distribution. In our case, this framework does not fit for two reasons: (1) we do not want our generator to fit our synthetic data's distribution (\textit{e.g.} a single object in the center of the image), which still lacks diversity compared to real data in spite of our efforts; (2) visual quality is not the sole concern of our problem, the main one being to accurately remove specularities. Therefore we propose a new framework following the GAN paradigm of Goodfellow et al.~\cite{goodfellow2016nips}, but differing in the discriminator network's role.

\subsubsection{Multi-class Adversarial Optimization}
The discriminator network's objective is to classify generated diffuse images from real ones but also from input images $I$, which show specularities. Instead of returning a single scalar value, it ouputs a tensor of three values, standing for the probabilities of the image belonging to either one of the three classes and which add up to $1$. For that we replaced the usual sigmoid activation function at the last layer by a softmax activation. We call $\mathcal{D}_{\theta_d}$ our discriminator network and $\Hat{D}=\mathcal{G}_{\theta_g}(I)$ a diffuse image generated by $\mathcal{G}_{\theta_g}$. The discriminator network is then optimized in an alternating manner with $\mathcal{G}_{\theta_g}$ to solve the adversarial min-max problem:
\begin{equation}
\begin{split}
\min_{\theta_g}\max_{\theta_d} & \,\,\sum_{i=1}^{3} \mathbb{E}_{x_i\sim p_{x_i}}\left\lbrace \log \mathcal{D}_{\theta_d}^{(i)}(x_i) \right\rbrace 
+ \sum_{i=1, j=1, i\neq j}^3\mathbb{E}_{x_j\sim p_{x_j}}\left\lbrace \log\left[ 1 -  \mathcal{D}_{\theta_d}^{(i)}(x_j) \right] \right\rbrace,
\end{split}
\label{eq:minmax}
\end{equation}
where:
\begin{itemize}
    \item $\mathcal{D}_{\theta_d}^{(i)}$ denotes the $i^{th}$ output of $\mathcal{D}_{\theta_d}$ and represents the probability of the image belonging to the class $C_i$.
    \item $x_i$ is an image drawn from the distribution $ p_{x_i}$ which corresponds to $C_i$. 
    \item $ C_i \in \left\lbrace C_1, C_2, C_3 \right\rbrace = \left\lbrace C_I, C_D, C_{\Hat{D}} \right\rbrace $ is one of the three classes.
\end{itemize}
Note that Eq.~\eqref{eq:minmax} depends on $\theta_g$ when $j=3$ and $x_j=\Hat{D}=\mathcal{G}_{\theta_g}(I)$. The idea behind this multi-class discriminator is that recognizing $D$ from $\hat{D}$ will ensure visual quality as in a classical GAN, while the classification between $I$ and $\hat{D}$ will compel the discriminator to find features related to the sole difference between them \textit{i.e.} specular highlights, thus ensuring accurate specularity removal.

\subsubsection{Architecture} An overview of the discriminator's architecture can be seen in figure~\ref{fig:overview} (bottom). It takes as input the image of size $256\times 256$ and decreases its resolution every other layer with convolutions of stride 2, going from $256\times 256$ to $16\times 16$, while the number of kernel filters increases every other layer, going from $64$ to $512$. Every convolutional layer is followed by a LeakyReLU activation and a batch normalization except at the last layer. We then apply a Flatten layer before a Dense layer to form the 3-dimensional vector.

\subsection{Loss Function}
\label{sec:loss}
To stabilize the learning and at the same time train an efficient and accurate network, we define our loss to be a combination of a content loss and our SSDS-specific adversarial loss:
\begin{equation}
\ell \left( \mathcal{G}_{\theta_g}(I), D \right) = \ell_{content} + \lambda\times \ell_{SSDS},
\label{eq:loss}
\end{equation}
where $\lambda$ is a regularization parameter to scale the two losses.

We use the Mean Square Error (MSE) as our content loss:
\vspace{-.3cm}
\begin{align}
    &\ell_{content} \left( \mathcal{G}_{\theta_g}(I), D \right) = \frac{1}{W H} \sum_{x=1}^W \sum_{y=1}^H \left( D_{x,y} - \mathcal{G}_{\theta_g}(I_{x,y}) \right)^2,
\end{align}
with $W$ and $H$ are respectively the width and the height of the input image. We explored other options, such as measuring the error on low layer feature maps of a discriminator network in order to extract relevant representations~\cite{ledig2016photo,seddik2019generative}, but it did not bring improvements on our rather simple data.

The discriminative part of the loss is defined on the probabilities of the discriminator and updates the weights of the generator via the gradient of:
\begin{equation}
\begin{split}
    \ell_{SSDS} = \log\left\lbrace D^{(3)}_{\theta_d}(\mathcal{G}_{\theta_g}(I))\right\rbrace 
    - \log\left\lbrace D^{(1)}_{\theta_d}(\mathcal{G}_{\theta_g}(I))\right\rbrace
    - \log\left\lbrace D^{(2)}_{\theta_d}(\mathcal{G}_{\theta_g}(I))\right\rbrace ,
\end{split}
    \label{eq:loss2}
\end{equation}
where, as a reminder, $3$, $2$ and $1$ correspond to $\hat{D}$, $D$ and $I$ respectively.
Compared to the adversarial loss of the GAN as formulated by Goodfellow et al.~\cite{goodfellow2016nips}, our multi-class adversarial loss has two more terms, which translate into more gradients for the back-propagation.

%-------------------------------------------------------------------------
\subsection{Training details}
We trained our networks from scratch simultaneously on an NVIDIA GeoForce GTX using the dataset described in \S\ref{sec:Dataset}. We trained the generator and the discriminator in an alternating manner with batches of size $16$. We scaled the range of the input images to $[0, 1]$. Our final model was trained for $30,000$ iterations at a learning rate of $2\cdot 10^{-4}$ and a decay of $0$ using the ADAM optimizer~\cite{kingma2014adam}. The generator and the discriminator were updated at each iteration to solve the adversarial min-max~\eqref{eq:minmax}. In addition, the generator was updated to solve the optimization problem~\eqref{eq:final_obj} via the gradient of the loss~\eqref{eq:loss}, with a regularization parameter set to $\lambda = 10^{-3}$. We implemented our models in Keras~\cite{chollet2015keras}.

%-------------------------------------------------------------------------
%===================================
% EXPERIMENTS
%==================================
\section{Experiments} \label{Results}
In this section, we evaluate our Specularity Removal Network quantitatively on synthetic data, for which we have ground-truth, and qualitatively on real data. We compare the performance of our method with three states-of-the-art separation methods: Tan et al.~\cite{tan2005separating} and Shen et al.~\cite{shen2008chromaticity}, which are model-based methods, and Shi et al.~\cite{shi2017learning} which is learning-based. The code for the hand-crafted methods are available on the authors' webpages and we downloaded the model of ~\cite{shi2017learning} on the author's GitHub\footnote{https://github.com/shi-jian/shapenet-intrinsics}. We also show the contribution of the multi-class adversarial loss in \S\ref{sec:multi-class} and discuss the limits in \S\ref{sec:limits}.
%------------------------------------------------------------------------
%===================================
% Comparison to SOTA and BASELINES
%==================================
\subsection{Evaluation on Synthetic Data}
\begin{table}[h]
\begin{tabular}{cc}
\includegraphics[scale=0.125, valign=m]{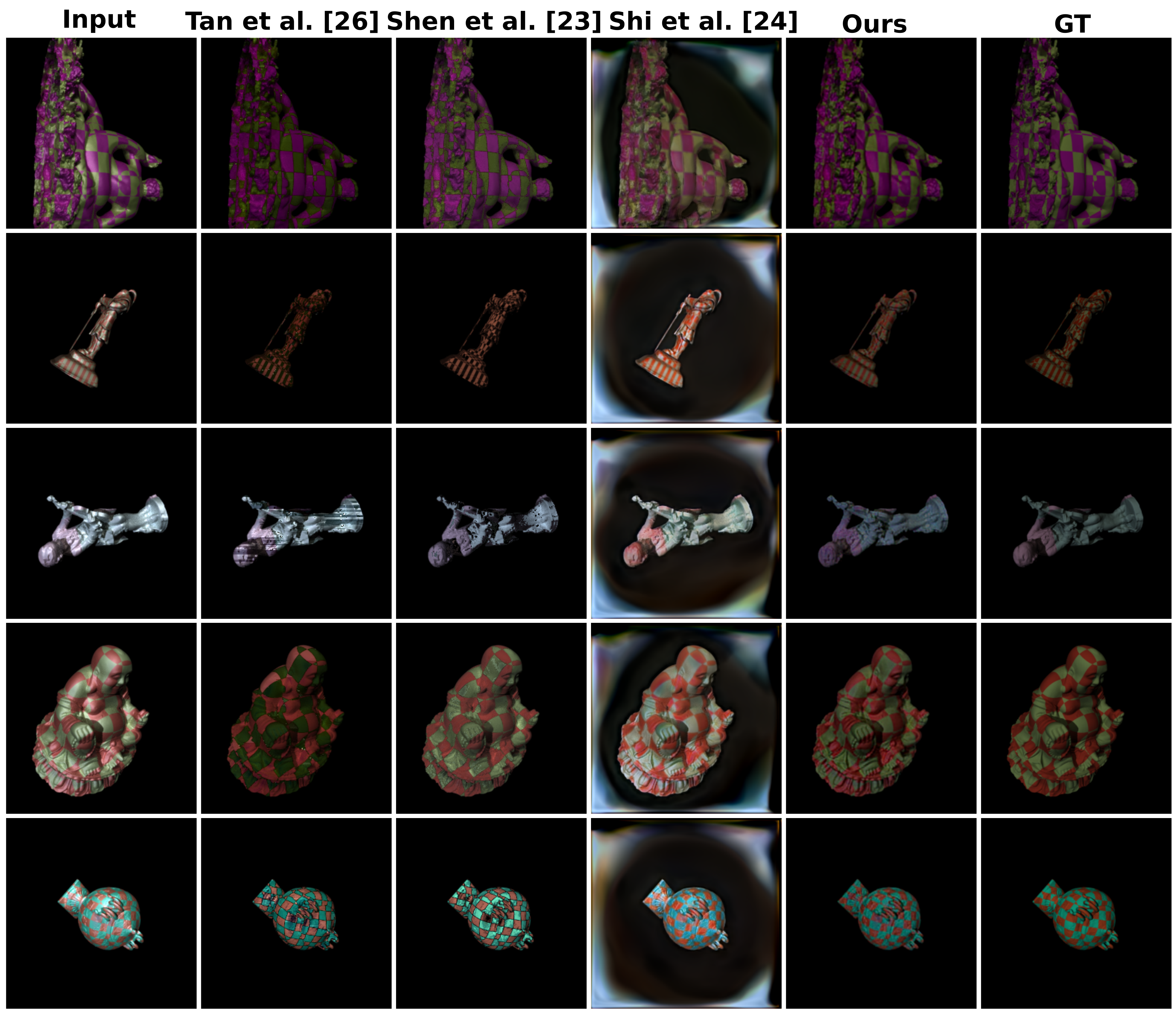}

\begin{tabular}{| l | c  c  c  c |}
\hline
Methods & L2 & DSSIM & NET & LIN-NET \\
\hline
\hline
Tan et al.~\cite{tan2005separating} & $0.020$ & $0.052$ & $ 0.408$ & $ 0.054$\\
Shen et al.~\cite{shen2008chromaticity} & $\underline{0.016} $ & $0.048 $  & $ \underline{0.380}$ & $0.052$\\
Shi et al.~\cite{shi2017learning} & $0.222$ & $0.410$ & $2.363$ & $0.313$\\
\hline
\hline
Baseline (AE) & $0.033$ & $\underline{0.046}$ & $\underline{0.336}$ & $\underline{0.044}$\\
Baseline (GAN) & $0.017$ & $0.059$ & $  0.474$ & $0.064 $ \\
Ours & $\textbf{0.014}$ & $\textbf{0.035}$ & $\textbf{0.271} $ &  $ \textbf{0.034}$\\
\hline
\end{tabular}

\end{tabular}
%\hspace{10cm}
\caption{(Left) Results of the \textbf{diffuse} component estimation task of the different methods on five synthetic test images. Best view in PDF. (Right) Quantitative results of the diffuse component estimation task in terms of the different metrics. The values represent the mean distance between the generated diffuse component and ground-truth over $1000$ randomly sampled synthetic test images. Baseline (AE) is a simple autoencoder without GAN training and Baseline refers to our method with binary classification (classical GAN framework).}
\label{table:results}
\end{table}

We provide qualitative and quantitative results regarding the estimation of the diffuse component of synthetic images in table~\ref{table:results}. Table~\ref{table:results} (left) shows the results of the diffuse component recovery with the different baseline methods and our method, on five images out of our test set of 1000 images. As can be noticed, our method's outcome is the perceptually closest to the sought diffuse components. Note that, the learning-based approach of Shi et al.~\cite{shi2017learning} shows reconstruction artifacts on the edges of the image. Their method actually has to work with a mask to segment the foreground object, which is limited as the mask is not provided in most scenarios. 

Table~\ref{table:results} (right) provides quantitative results of the different methods and two Baseline methods (AE for autoencoder and GAN for classical GAN) averaged on the test set. We use two standard metrics, namely L2 and DSSIM \cite{hore2010image}, and also consider recently developed perceptual metrics based on computing the similarity of two images in the feature space of a neural network. Indeed, Zhang et al.~\cite{zhang2018unreasonable} showed that these metrics provide an embedding of images which agrees surprisingly well with human judgment. In particular, we consider the metric corresponding to the Squeeze network~\cite{iandola2016squeezenet} (denoted NET) and its linear version as proposed by~\cite{zhang2018unreasonable} (denoted LIN-NET). This quantitative comparison agrees with the perceptual analysis as our method outperforms all the methods by a large margin on all the metrics. Specifically, it outperforms the Baseline method which consists in considering a binary classification for the discriminator rather than the proposed multi-class classification.

%Our method being based on a network that is trained on synthetic images, the fact that it outperforms the other methods on these images is quite natural. However, we see in the following section that even though we used synthetic data for training, our method gives convincing results on real images.

\subsection{Evaluation on Real Data}
\label{sec:res_real}
\begin{figure} [!h]
\centering
	\includegraphics [width = \linewidth]{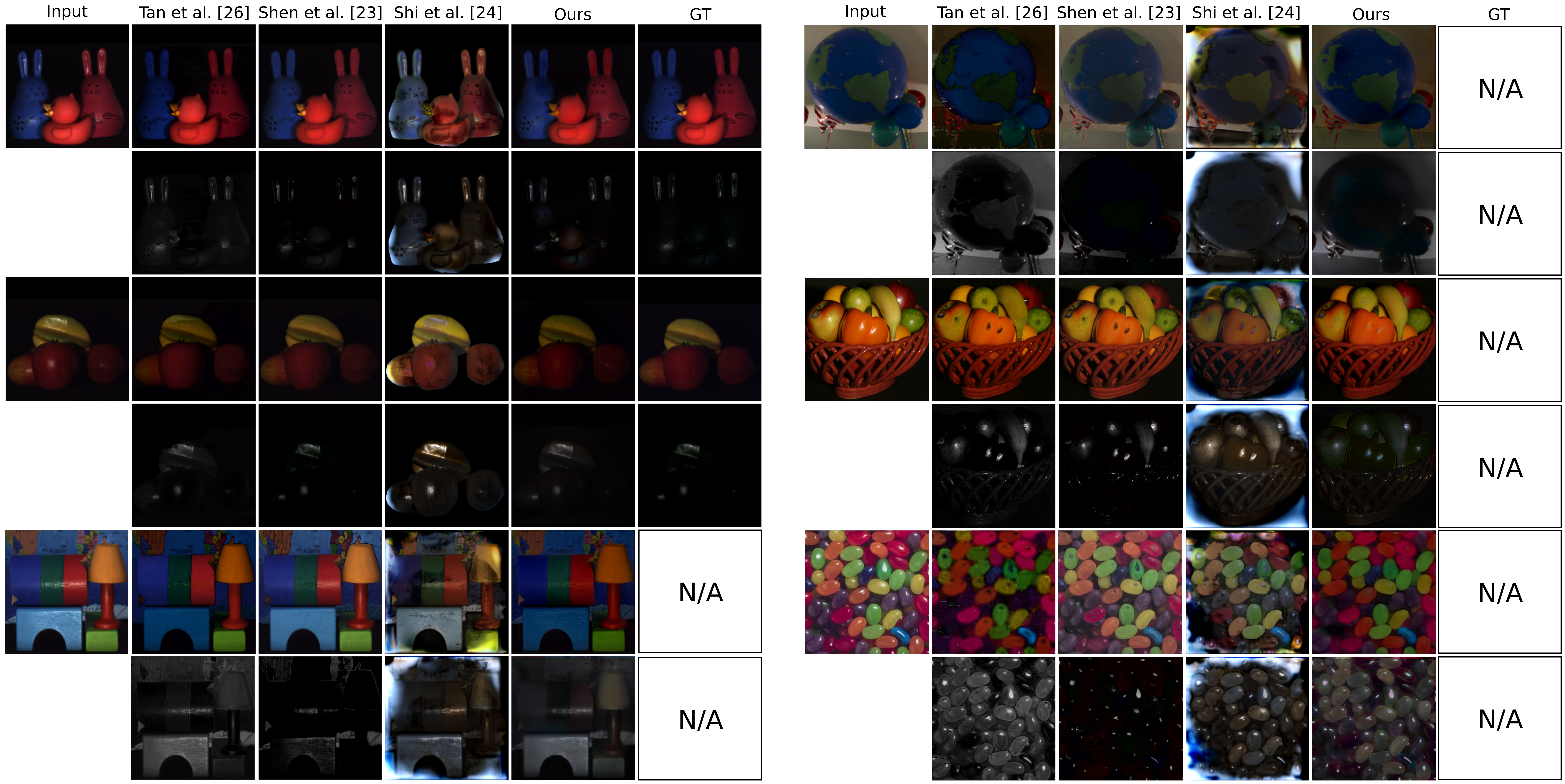}
\caption{\label{fig:res_real}
Results and comparison of our method on real images. The input image is on the left and for each image, the top row is the diffuse component and the bottom row is the specular component. Ground-truths are provided on the right when available. Our baselines include Tan et al.~\cite{tan2005separating}, Shen et al.~\cite{shen2008chromaticity} and Shi et al.~\cite{shi2017learning}. Our specular component is obtained by subtracting our estimated diffuse component to the input image. Best view in PDF.}
\end{figure}
Our network being trained on synthetic images, the fact that it outperforms the other methods on synthetic data can seem natural. However, in this section we evaluate our method on real images and show that it performs consistenly. The results can be seen in figure~\ref{fig:res_real}. Our network performs well on a wide range of images, from images resembling our training data with a black background to complex scenes such as the wooden objects and the earth balloon. This attests of a better generalization from our Specularity Removal Network, while artifacts on the edges and on the object are still visible in the results of Shi et al.~\cite{shi2017learning}. Our discriminative network constrains the generator to understand the distribution of specularities and help it remove them from the image by training themselves to differentiate $I$ from $D$. Visually, our method consistenly outperforms \cite{shi2017learning}.

In visual comparison to hand-crafted methods, we perform slightly worse than Shen et al.~\cite{shen2008chromaticity} on the first and the second images (animals and fruits) and slightly worse than Tan et al.~\cite{tan2005separating} on the first and third examples (animals and wooden objects). This can be explained by the fact that these images were taken in laboratory conditions and fit perfectly the hypothesis of the Dichromatic Reflection Model \cite{shafer}, on which hand-crafted models are built. However, albeit subjective since there is no metric to evaluate specularity removal, our method performs best on the other examples, showing its consistency. On the fifth image (the fruit basket), we can clearly see that purely specular (saturated) pixels lose too much energy with hand-crafted methods because they don't show any diffuse color underneath, while our method consistently begins to inpaint them. The same goes for the last image (beans) where our method does not leave holes like the others. In summary, our method separates the reflection components with the best consistency compared to the state-of-the-art. 

\subsection{Contribution of our Multi-Class GAN} \label{sec:multi-class}
\begin{figure}[h]
    \centering
    \includegraphics[width=0.8\linewidth]{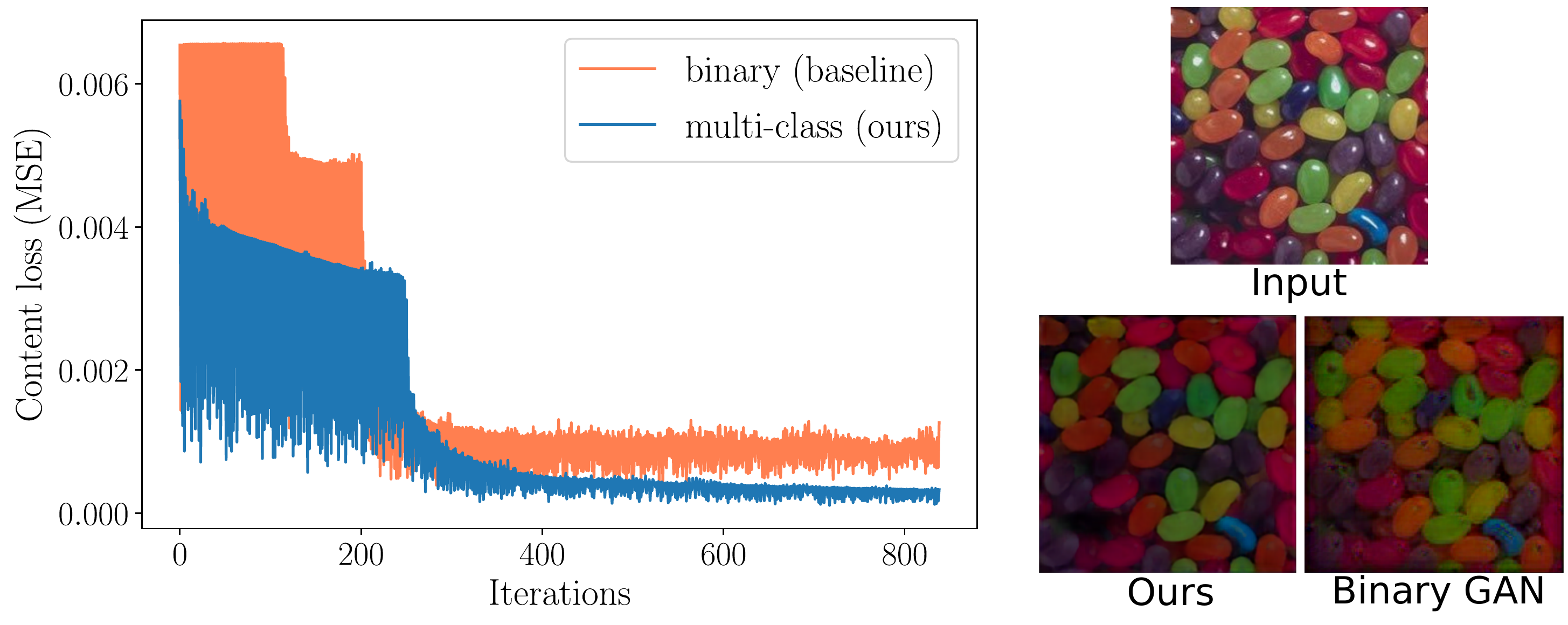}
    \caption{(Left) Learning curves of a classical GAN and our multi-class GAN. (Right) Generated diffuse components.}
    \label{fig:courbe}
\end{figure}

Figure~\ref{fig:courbe} shows learning curves for our method and for a classical binary GAN with the exact same parameters. The amplitudes of the oscillations of both curves easily tell us that the multi-class adversarial loss allows for a more stable learning, while instability is common to adversarial frameworks. It also shows that convergence comes faster and is more accurate, which is visible in the images generated by the two methods (right of figure~\ref{fig:courbe}).

\subsection{Limits} \label{sec:limits}
Our network is of course not perfect and can have failure cases in a real life application. First, as mentioned before it does not handle mirror surfaces, which would require a completely different definition of the problem (different data and priors). Furthermore, despite our efforts (white objects and multi-class discriminators), the network still tends to darken the images. This is visible in the earth balloon example of figure~\ref{fig:res_real} and shows our network still misses a step of generalization. This can be explained by the simplicity of our data, especially the lack of a background which would normally provide context to work on. Not any random background however, which would have been easy to add, but a true background containing illumination information. We tried to add a white background to help the network discern white material from white specular highlights but did not see any noticeable improvement. We also tried the patchGAN formulation of the GAN framework without noticeable changes.

\section{Conclusion} 
\label{conclusion}

% We show that generating the specular component is not the ideal solution to get the diffuse part, which motivates our choice since it is the part that interests us. Indeed, we observe in Figure [] that subtracting the specular reflection to the image generally leaves the specular highlights intact or on the contrary creates holes because the intensity and color are wrongly estimated, whereas the diffuse component has perceptual cues from purely diffuse pixels at its disposition.

We have proposed a new method to separate diffuse and specular reflections based on a learning approach, which can better infer the complex relations between the object, the lighting condition and the image than existing approaches. Our method takes advantage of synthetic data generation which allows us to easily obtain a large amount of labeled data. We generated our own training set and augmented our data in such a way to account for difficult cases encountered in real scenarios, in order to help the network better generalize to these situations. We also trained our model in a GAN framework adapted to reflection component separation. For that, we defined a new multi-class adversarial loss, which helps the training process by providing more gradients and more precise features. This results in a model that removes specularity from a single image, without any assumption made about the scene. We evaluated our model on both synthetic and real data. Our method outperforms the state-of-the-art in consistency across various scenes.

In future work, we would like to investigate the temporal coherence for live applications in a continuous video stream. Our Specularity Removal Network is not perfect and might not be entirely consistent from one frame to the next. Our network would also greatly benefit from data with more complex scenes in the training set.
%
% the environments 'definition', 'lemma', 'proposition', 'corollary',
% 'remark', and 'example' are defined in the LLNCS documentclass as well.
%

\bibliographystyle{splncs04}
\bibliography{mybibliography}

\begin{thebibliography}{10}
\providecommand{\url}[1]{\texttt{#1}}
\providecommand{\urlprefix}{URL }
\providecommand{\doi}[1]{https://doi.org/#1}

\bibitem{adelson1996perception}
Adelson, E.H., Pentland, A.P.: The perception of shading and reflectance.
  Perception as Bayesian inference pp. 409--423 (1996)

\bibitem{an2015fast}
An, D., Suo, J., Ji, X., Wang, H., Dai, Q.: Fast and high quality highlight
  removal from a single image. arXiv preprint arXiv:1512.00237  (2015)

\bibitem{chollet2015keras}
Chollet, F., et~al.: Keras (2015)

\bibitem{funke2018generative}
Funke, I., Bodenstedt, S., Riediger, C., Weitz, J., Speidel, S.: Generative
  adversarial networks for specular highlight removal in endoscopic images. In:
  Medical Imaging 2018: Image-Guided Procedures, Robotic Interventions, and
  Modeling. vol. 10576, p. 1057604. International Society for Optics and
  Photonics (2018)

\bibitem{goodfellow2016nips}
Goodfellow, I.: Nips 2016 tutorial: Generative adversarial networks. arXiv
  preprint arXiv:1701.00160  (2016)

\bibitem{hara2003determining}
Hara, K., Nishino, K., Ikeuchi, K.: Determining reflectance and light position
  from a single image without distant illumination assumption. In: null.
  p.~560. IEEE (2003)

\bibitem{hore2010image}
Hore, A., Ziou, D.: Image quality metrics: Psnr vs. ssim. In: Pattern
  recognition (icpr), 2010 20th international conference on. pp. 2366--2369.
  IEEE (2010)

\bibitem{iandola2016squeezenet}
Iandola, F.N., Han, S., Moskewicz, M.W., Ashraf, K., Dally, W.J., Keutzer, K.:
  Squeezenet: Alexnet-level accuracy with 50x fewer parameters and< 0.5 mb
  model size. arXiv preprint arXiv:1602.07360  (2016)

\bibitem{jachnik2012real}
Jachnik, J., Newcombe, R.A., Davison, A.J.: Real-time surface light-field
  capture for augmentation of planar specular surfaces. In: Mixed and Augmented
  Reality (ISMAR), 2012 IEEE International Symposium on. pp. 91--97. IEEE
  (2012)

\bibitem{kingma2014adam}
Kingma, D.P., Ba, J.: Adam: A method for stochastic optimization. arXiv
  preprint arXiv:1412.6980  (2014)

\bibitem{klinker1988measurement}
Klinker, G.J., Shafer, S.A., Kanade, T.: The measurement of highlights in color
  images. International Journal of Computer Vision  \textbf{2}(1),  7--32
  (1988)

\bibitem{lambert1760photometria}
Lambert, J.H.: Photometria sive de mensura et gradibus luminis, colorum et
  umbrae. Klett (1760)

\bibitem{ledig2016photo}
Ledig, C., Theis, L., Husz{\'a}r, F., Caballero, J., Cunningham, A., Acosta,
  A., Aitken, A., Tejani, A., Totz, J., Wang, Z., et~al.: Photo-realistic
  single image super-resolution using a generative adversarial network. arXiv
  preprint  (2016)

\bibitem{lin1999estimation}
Lin, S., Lee, S.W.: Estimation of diffuse and specular appearance. In: Computer
  Vision, 1999. The Proceedings of the Seventh IEEE International Conference
  on. vol.~2, pp. 855--860. IEEE (1999)

\bibitem{lin2002diffuse}
Lin, S., Li, Y., Kang, S.B., Tong, X., Shum, H.Y.: Diffuse-specular separation
  and depth recovery from image sequences. In: European conference on computer
  vision. pp. 210--224. Springer (2002)

\bibitem{lin2001separation}
Lin, S., Shum, H.Y.: Separation of diffuse and specular reflection in color
  images. In: Computer Vision and Pattern Recognition, 2001. CVPR 2001.
  Proceedings of the 2001 IEEE Computer Society Conference on. vol.~1,
  pp.~I--I. IEEE (2001)

\bibitem{meka2018live}
Meka, A., Maximov, M., Zollhoefer, M., Chatterjee, A., Richardt, C., Theobalt,
  C.: Live intrinsic material estimation. arXiv preprint arXiv:1801.01075
  (2018)

\bibitem{nayar1997separation}
Nayar, S.K., Fang, X.S., Boult, T.: Separation of reflection components using
  color and polarization. International Journal of Computer Vision
  \textbf{21}(3),  163--186 (1997)

\bibitem{ronneberger2015u}
Ronneberger, O., Fischer, P., Brox, T.: U-net: Convolutional networks for
  biomedical image segmentation. In: International Conference on Medical image
  computing and computer-assisted intervention. pp. 234--241. Springer (2015)

\bibitem{seddik2019generative}
Seddik, M.E.A., Tamaazousti, M., Lin, J.: Generative collaborative networks for
  single image super-resolution. arXiv:1902.10467  (2019)

\bibitem{shafer}
Shafer, S.A.: Using color to separate reflection components. COLOR Research \&
  Application  \textbf{10}(4),  210--218 (1985)

\bibitem{shen2009simple}
Shen, H.L., Cai, Q.Y.: Simple and efficient method for specularity removal in
  an image. Applied optics  \textbf{48}(14),  2711--2719 (2009)

\bibitem{shen2008chromaticity}
Shen, H.L., Zhang, H.G., Shao, S.J., Xin, J.H.: Chromaticity-based separation
  of reflection components in a single image. Pattern Recognition
  \textbf{41}(8),  2461--2469 (2008)

\bibitem{shi2017learning}
Shi, J., Dong, Y., Su, H., Stella, X.Y.: Learning non-lambertian object
  intrinsics across shapenet categories. In: Computer Vision and Pattern
  Recognition (CVPR), 2017 IEEE Conference on. pp. 5844--5853. IEEE (2017)

\bibitem{shi2016real}
Shi, W., Caballero, J., Husz{\'a}r, F., Totz, J., Aitken, A.P., Bishop, R.,
  Rueckert, D., Wang, Z.: Real-time single image and video super-resolution
  using an efficient sub-pixel convolutional neural network. In: Proceedings of
  the IEEE Conference on Computer Vision and Pattern Recognition. pp.
  1874--1883 (2016)

\bibitem{tan2005separating}
Tan, R.T., Ikeuchi, K.: Separating reflection components of textured surfaces
  using a single image. IEEE transactions on pattern analysis and machine
  intelligence  \textbf{27}(2),  178--193 (2005)

\bibitem{weiss2001deriving}
Weiss, Y.: Deriving intrinsic images from image sequences. In: Computer Vision,
  2001. ICCV 2001. Proceedings. Eighth IEEE International Conference on.
  vol.~2, pp. 68--75. IEEE (2001)

\bibitem{wolff1991constraining}
Wolff, L.B., Boult, T.E.: Constraining object features using a polarization
  reflectance model. IEEE Transactions on Pattern Analysis and Machine
  Intelligence  \textbf{13}(7),  635--657 (1991)

\bibitem{zhang2018unreasonable}
Zhang, R., Isola, P., Efros, A.A., Shechtman, E., Wang, O.: The unreasonable
  effectiveness of deep features as a perceptual metric. arXiv preprint  (2018)

\end{thebibliography}

\end{document}